\documentclass[conference]{IEEEtran}
\IEEEoverridecommandlockouts
\usepackage{cite}
\usepackage{amsmath,amssymb,amsfonts}
\usepackage{algorithmic}
\usepackage{graphicx}
\usepackage{textcomp}
\usepackage[table]{xcolor}
\def\BibTeX{{\rm B\kern-.05em{\sc i\kern-.025em b}\kern-.08em
    T\kern-.1667em\lower.7ex\hbox{E}\kern-.125emX}}

\usepackage{booktabs}
\usepackage{subcaption}
\usepackage{multirow}
\graphicspath{{./images/}}

\definecolor{Gray}{gray}{0.95}
\newcolumntype{g}{>{\columncolor{Gray}}c}    
\begin{document}

\title{Kernel density estimation-based sampling for neural network classification
\thanks{© 20XX IEEE.  Personal use of this material is permitted.  Permission from IEEE must be obtained for all other uses, in any current or future media, including reprinting/republishing this material for advertising or promotional purposes, creating new collective works, for resale or redistribution to servers or lists, or reuse of any copyrighted component of this work in other works.}
}

\author{\IEEEauthorblockN{Firuz Kamalov}
\IEEEauthorblockA{\textit{Department of Electrical Engineering} \\
\textit{Canadian University Dubai}\\
Dubai, UAE\\
firuz@cud.ac.ae}
\and
\IEEEauthorblockN{Ashraf Elnagar}
\IEEEauthorblockA{\textit{Department of Computer Science} \\
\textit{University of Sharjah}\\
Sharjah, UAE \\
ashraf@sharjah.ac.ae}
}

\maketitle

\begin{abstract}
Imbalanced data occurs in a wide range of scenarios. The skewed distribution of the target variable elicits bias in machine learning algorithms. One of the popular methods to combat imbalanced data is to artificially balance the data through resampling. In this paper, we compare the efficacy of a recently proposed kernel density estimation (KDE) sampling technique in the context of artificial neural networks. We benchmark the KDE sampling method against two base sampling techniques and perform comparative experiments using 8 datasets and 3 neural networks architectures. The results show that KDE sampling produces the best performance on 6 out of 8 datasets. However, it must be used with caution on image datasets. We conclude that KDE sampling is capable of  significantly improving the performance of neural networks.
\end{abstract}


\section{Introduction}
Imbalanced data refers to skewed distribution of the target variable in the dataset. It occurs in a range of fields including medical diagnostics, cybersecurity, fraud detection, text categorization, and many others. Imbalanced data can cause bias in machine learning classifiers \cite{Thabtah}. Since the objective of a classifier is to minimize the overall error rate it focuses on correctly classifying the majority class subset at the expense of the minority class. However, the minority class data is often of more importance than the majority class set. For instance, although only a minority of credit card transactions data is fraudulent identifying such transactions is critical.
A commonly employed approach to combat imbalanced data is resampling the original data. Concretely, oversampling the minority set points is often used to achieve a balanced dataset.
In this paper, we analyze the effectiveness of a recently proposed sampling technique based on KDE in the context of artificial neural networks. 

The KDE sampling is carried out by first estimating the underlying distribution of the existing points and then generating the new minority points from the estimated distribution. As a result, KDE offers an effective and natural way to sample new points. The KDE sampling was originally proposed in \cite{Kamalov1} where the authors demonstrated the effectiveness of the sampling method on k-nearest neighbors, support vector machines, and multilayer perceptron classifiers. Our goal is to conduct a more in-depth analysis of the performance of the KDE sampling with neural networks.

We employ neural networks of various depths to analyze the performance of the KDE sampling. We train and test the neural networks on 8 different datasets from a range of applications. We also employ two standard sampling approaches as benchmarks against the KDE sampling. The results demonstrate that the KDE sampling is effective in the majority of cases achieving the highest overall $F_1$-scores on 6 out of 8 datasets. We conclude that the KDE method can be a valuable tool in dealing with imbalanced data when using neural network classifiers.

\section{Literature Survey}

Imbalanced data occurs in a range of machine learning applications. In \cite{Fotouhi}, the authors study various sampling techniques in the context of cancer data. In \cite{Zhang}, the authors use generative adversarial networks (GAN) to balance machinery vibrations data. GANs are applied to train and generate artificial readings of faulty machinery states. Sampling techniques are evaluated in the context of software detection in \cite{Malhotra}. The authors of the study compare the performance of sampling and cost sensitive approaches to imbalanced learning using 12 NASA datasets. 
Oversampling is used to balance traffic incident data in \cite{Parsa}. The authors use SMOTE to preprocess highway traffic dat to develop an accident predicting intelligent model.

There exists a number of sampling approaches to deal with imbalanced data. A survey of the existing sampling methods can be found in \cite{Leevy}. Sampling approaches can be grouped into two categories: undersampling and oversampling.
Undersampling involves sampling from the majority subset to achieve the same size as the minority subset. 
In \cite{Lin}, the authors propose two undersampling methods where the majority points are divided into the same number of clusters as the number of the minority points. In the first approach, the new majority points are taken as the centers of the clusters whereas in the second approach the new majority points are taken as the nearest neighbors of the clusters. 
Oversampling involves generating new minority points to achieve the same number as the majority set. One of the popular oversampling methods SMOTE creates new points along the line segment joining a pair of neighboring minority points \cite{Fernandez}. More recently, a Gamma distribution-based oversampling approach was introduced in \cite{Kamalov2}. The authors generate the new points along the straight line between two existing minority points using the Gamma distribution. Oversampling for multiclass data is explored in \cite{Krawczyk}. The authors propose to generate new data samples in the regions of space with sparse mutual class distribution.

Sampling has been used to address the issue of imbalanced data in the context of neural networks. In \cite{Shen}, the authors employed the SMOTE algorithm to balance the data before using backpropagation neural network to analyze credit card data. In \cite{Johnson}, the author evaluate the performance of ROS and RUS with deep learning in the context of Medicare fraud detection.


\section{Methodology}
\subsection{Sampling methods}
KDE is a well-known nonparametric density estimator used in number of applications \cite{Kamalov3, Lee}.
The KDE-based sampling approach is based on estimating the underlying distribution of the minority data. Given a set of existing minority points $x_1, x_2, ..., x_n$ the underlying density distribution ${f}$ can be approximated by 
\begin{equation}
\tilde{f}(x) = \frac{1}{n}\sum_{i=1}^n K_h(x-x_i) ,
\end{equation}
where $K$ is the kernel function, $h$ is the bandwidth parameter, and $K_h(t) = \frac{1}{h}K(\frac{t}{h})$.   The optimal bandwidth value can be determined numerically through cross-validation. It is done by applying a grid search method to find the value of $h$ that minimizes the sample mean integrated square error:
\begin{equation}
\mbox{MISE}_n(h) = \frac{1}{n} \sum_{i=1}^n (\tilde{f}(x_i)-f(x_i))^2.
\end{equation}
Default values of $h$ exist under certain assumptions about the underlying density function $f$. A standard approach to calculating the optimal value of $h$ is given by Scott's rule:
\begin{equation}\label{h_val}
h  = n^{-\frac{1}{5}}\cdot s,
\end{equation}
where $s$ is the sample  standard deviation. 

We benchmark the performance of KDE sampling against two standard sampling techniques: random oversampling (ROS) and random undersampling (RUS). The ROS technique balances the data by randomly selecting with replacement from the existing minority points. The main disadvantage of the ROS is overfitting. Since the ROS selects points with replacement it essentially replicates the minority set multiple times creating points of high concentration. The RUS technique randomly selects a subset of the majority set to match the size of the minority set. The main disadvantage of the RUS is the loss of information since only a portion of the available data is utilized.

\subsection{Experimental Data}
We employ 8 different imbalanced datasets in our numerical experiments. The datasets are available through the imblearn library \cite{Lemaitre} or the UCI repository \cite{Dua}. The details of the datasets are provided in Table \ref{info}.  The imbalance ratio in the datasets ranges from $9.3:1$ to $42:1$. The datasets include a diverse areas of application including medical, meteorology, computer vision, and others. 
\begin{table}[!htb]
\centering
\caption{Details of the experimental datasets.}
\label{info}
\begin{tabular}{lrrrr}
\toprule
Name &      Ratio &     \#Samples &   \#Features \\
\midrule
abalone &        9.7:1 &   4177 &   10 \\
letter\_img &    26:1 &  20,000 &   16\\
libras\_move &   14:1 &    360 &   90 \\
mammography &  42:1 &  11183 &    6 \\
ozone\_level &      34:1 &   2536 &   72 \\
satimage &    9.3:1 &   6435 &   36 \\
spectrometer &    11:1 &     531 &   93 \\
wine\_quality &    26:1 &   4898 &   11 \\ 
\bottomrule
\end{tabular}
\end{table}

\subsection{Neural network classifiers}
The performances of the KDE sampling and the benchmark methods are tested on 3 different neural network architectures with 1, 2, and 3 hidden layers. Deeper networks would be redundant due to the relatively small size of the datasets used in our experiments. The details of the three neural networks are presented in Table \ref{nn_details}.

\begin{table}[!htb]
\centering
\caption{Details of the neural networks used in the experiments.}
\label{nn_details}
\begin{tabular}{lrrrrr}
\toprule
Classifier &      Layer 1 &    Layer 2 &  Layer 3 & Activation & Optimizer\\
\midrule
MLP-1 & 64 & - & - & \multirow{3}{6em}{Relu, sigmoid (output layer)} & \multirow{3}{4em}{RMSprop}\\
MLP-2 & 32 & 8 & - &\\
MLP-3 & 64 & 32&  4 &\\
\bottomrule
\end{tabular}
\end{table}

As noted earlier, the classifier performance on the minority labeled data is often of far more importance than the majority data. For instance, in medical diagnostics it is crucial to identify the positive cases even if occasionally producing false positive results. Similarly, in network intrusion detection catching malicious attacks is the primary concern even if they comprise only a small portion of network traffic. Therefore, the accuracy on the minority set must play an important role when measuring the performance of the sampling methods.
In our experiments, the performance of the neural networks is measured with the macro average $F_1$-score which is well suited for imbalanced datasets. The macro average $F_1$-score is calculated as an equally weighted average $F_1$-score of the majority and minority labeled data
\begin{equation}
F_1^{\mbox{\tiny{macro}}} =  \frac{F_1^{\mbox{\tiny{major}}} + F_1^{\mbox{\tiny{minor}}}}{2}.
\end{equation}
As a result, it properly reflects the performance on the minority set.

\section{Results}
\begin{figure}[h!]
\center
\includegraphics[width=0.5\textwidth]{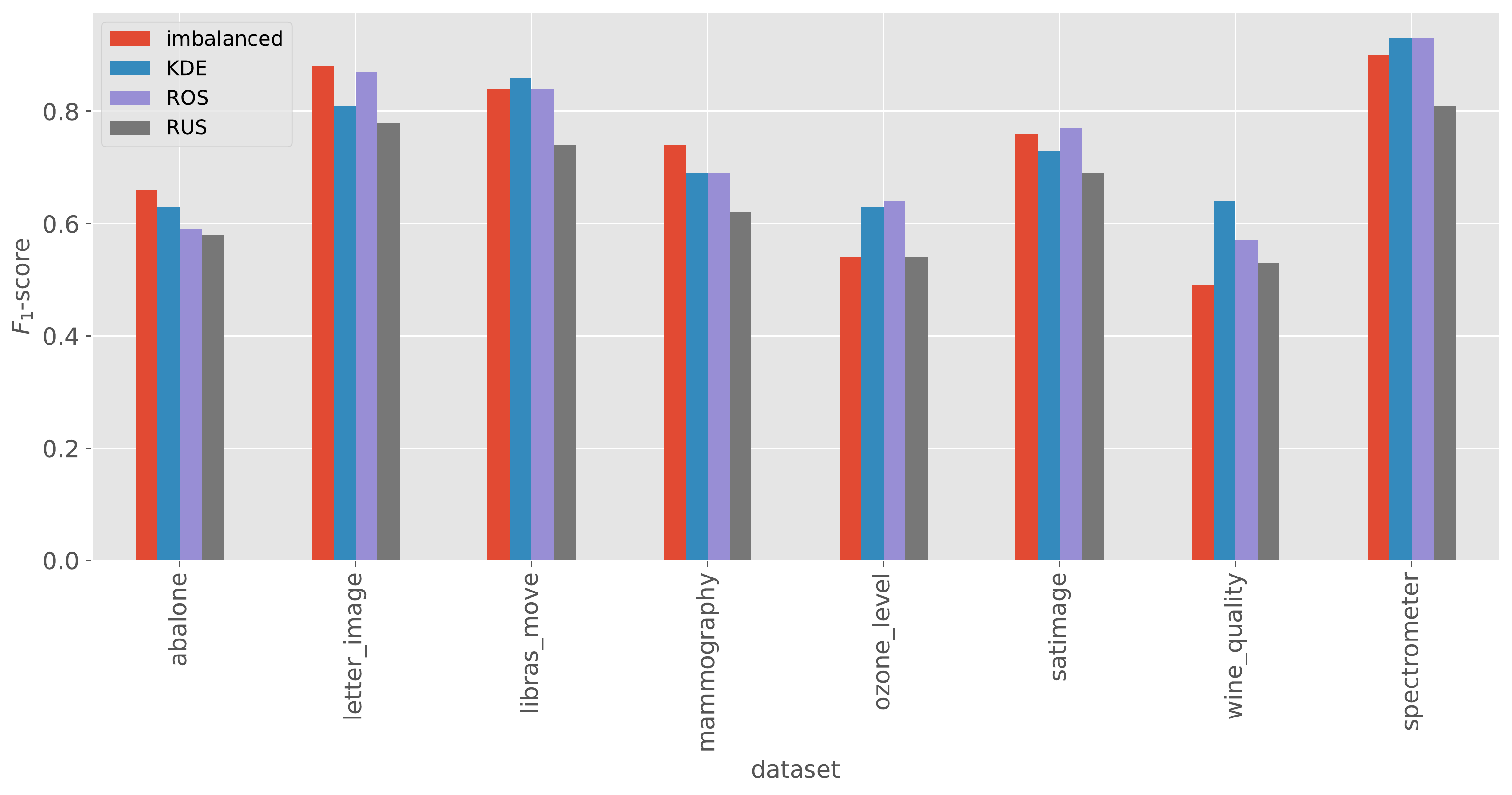}
\caption{Macro average $F_1$-score for MLP with 1 hidden layers.}
\label{ann1}
\end{figure}

\begin{figure}[h!]
\center
\includegraphics[width=0.5\textwidth]{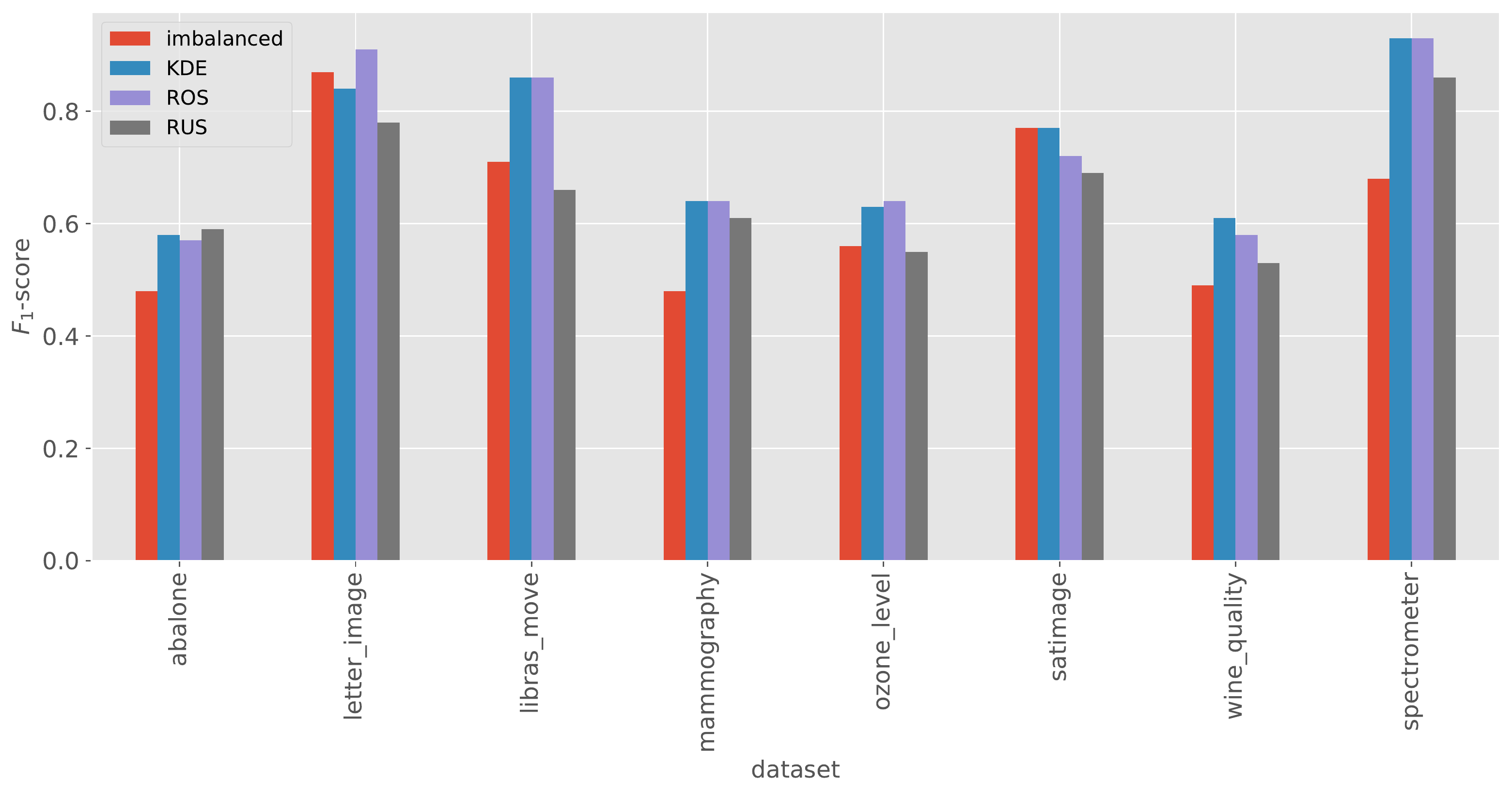}
\caption{Macro average $F_1$-score for MLP with 2 hidden layers.}
\label{ann2}
\end{figure}

\begin{figure}[h!]
\center
\includegraphics[width=0.5\textwidth]{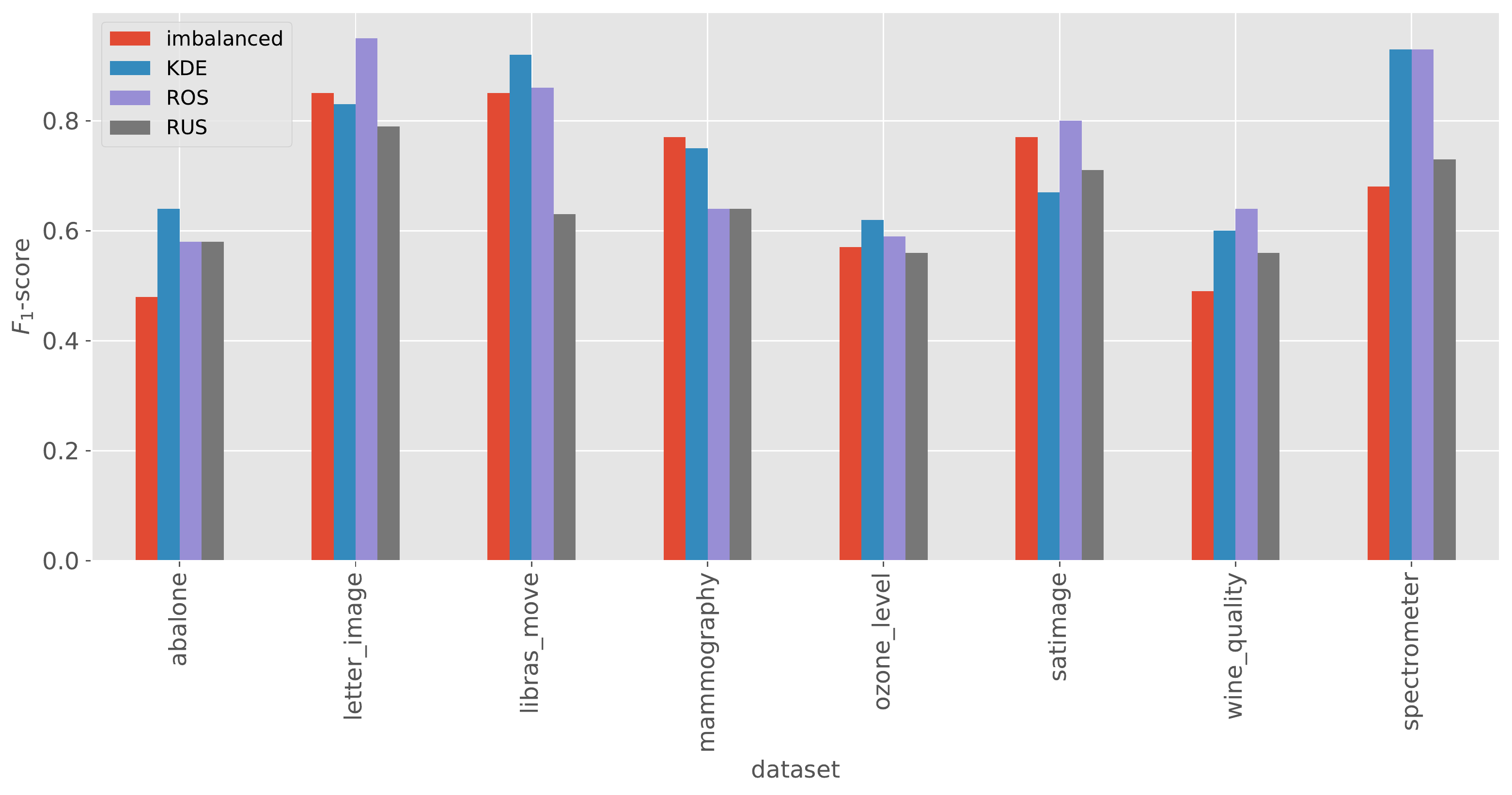}
\caption{Macro average $F_1$-score for MLP with 3 hidden layers.}
\label{ann3}
\end{figure}

The detailed results of the experiments are provided in Figures \ref{ann1}-\ref{ann3}. As can be seen from the figures, the KDE-based approach outperforms the benchmark methods. 
The KDE sampling performs particularly well on the \textit{abalone, libras\_move, mammography} and \textit{wine\_quality} datasets. On the other hand, it does not perform well on image recognition tasks and the corresponding datasets \textit{letter\_image} and \textit{satimage}.

In the experiments using a 1-layer neural network, the KDE sampling method performs well in all but two datasets (Figure \ref{ann1}).
It achieves either the best or nearly the best results. The only datasets with poor performance are the image datasets \textit{letter\_image} and \textit{satimage}. In the  \textit{abalone} dataset, the KDE sampling outperforms the benchmarks but lags behind the original imbalanced dataset.
In the experiments using a 2-layer neural network, the KDE sampling method performs well in all but one dataset (Figure \ref{ann2}). As above, it achieves either the best or nearly the best results. The only dataset with poor performance is the image dataset \textit{letter\_image}. Note that the performance is good on the second image dataset \textit{satimage}. 
The experiments with a 3-layer neural network yield mixed results (Figure \ref{ann3}). The KDE sampling achieves the best $F_1$-score on the three datasets \textit{abalone, libras\_move} and \textit{ozone\_level}. It is tied with the ROS method on the \textit{spectrometer}. 


To obtain the overall comparison of the sampling methods we averaged the $F_1$-scores for each dataset over the three neural networks. 
A summary of the overall results is presented in Table \ref{summary}. As shown in the table, the KDE sampling achieves the best overall score on 6 out of 8 datasets and tied on 1 dataset. Furthermore, we observe that the KDE sampling method performs poorly on the image datasets \textit{letter\_image} and \textit{satimage}. The poor performance on the image datasets is not surprising as the process of generating new image samples must take into account the internal structure of image files. Small changes in the values of an image file can easily result in a nonsensical image. As a result the new samples would be unhelpful.

\begin{table}[!htb]
\centering
\caption{Summary of averaged the $F_1$-scores for each dataset over the three neural networks.}
\label{summary}
\begin{tabular}{lrrrr}
\toprule
Dataset &  imbalanced &  KDE &  ROS &  RUS \\
\midrule
abalone &  0.5400 &      0.6167 &     0.5800 &       0.5833 \\
{letter\_im} &   0.8667 & 0.8267 &  0.9100 &       0.7833 \\
{libras} &   0.8000 &    0.8800 & 0.8533 & 0.6767 \\
{mammo} &     0.6633 &  0.6933 & 0.6567 &  0.6233 \\
{ozone} &  0.5567 &  0.6267 &    0.6233 &0.5500 \\ 
{satimage} &  0.7667 & 0.7233 &      0.7633 &  0.6967 \\   
{wine} &    0.4900 &     0.6167 &    0.5967 &     0.5400 \\   
{spectrometer} &  0.7533 & 0.9300  &  0.9300    &0.8000  \\    
\bottomrule
\end{tabular}
\end{table}

The KDE sampling algorithm outperforms the benchmark methods due to a more natural approach to sampling. The RUS method reduces the size of the majority set thereby losing valuable information contained in the original set. The ROS method essentially replicates the original minority set multiple times leading to overfitting. The KDE method avoids the above problems associated with the ROS and RUS methods. It samples from the estimated distribution providing an organic sampling procedure.

It is worth noting that, in theory, ROS and RUS are computationally more efficient than KDE sampling. Since ROS and RUS  perform simple random sampling of the original dataset the associated algorithmic complexity is very small. The KDE sampling requires estimation of the underlying distribution prior to sampling which requires additional time. Modern implementations of the KDE have linear complexity $O(n)$.
\section{Conclusion}
Imbalanced data occurs in a range of machine learning applications. Sampling is one of the effective approaches to deal with imbalanced data. 
In this paper, we analyzed the performance of a recently proposed KDE-based sampling method in the context of neural network classifiers. The performance of KDE sampling was compared against two standard benchmark sampling methods as well as the original imbalanced dataset. The results show that KDE sampling  achieves the best overall score in 6 of the 8 datasets used in the experiments. However, KDE sampling performs poorly on image data. The underperformance is explained by the internal structure on image files that require a more nuanced approach. We concluded that the KDE sampling offers a viable tool to deal with imbalanced data when using neural network classifiers.

\end{document}